# EnhanceNet: Single Image Super-Resolution Through Automated Texture Synthesis

Mehdi S. M. Sajjadi   Bernhard Schölkopf   Michael Hirsch

Max Planck Institute for Intelligent Systems
Spemanstr. 34, 72076 Tübingen, Germany

{msajjadi,bs,mhirsch}@tue.mpg.de

## Abstract

*Single image super-resolution is the task of inferring a high-resolution image from a single low-resolution input. Traditionally, the performance of algorithms for this task is measured using pixel-wise reconstruction measures such as peak signal-to-noise ratio (PSNR) which have been shown to correlate poorly with the human perception of image quality. As a result, algorithms minimizing these metrics tend to produce over-smoothed images that lack high-frequency textures and do not look natural despite yielding high PSNR values.*

*We propose a novel application of automated texture synthesis in combination with a perceptual loss focusing on creating realistic textures rather than optimizing for a pixel-accurate reproduction of ground truth images during training. By using feed-forward fully convolutional neural networks in an adversarial training setting, we achieve a significant boost in image quality at high magnification ratios. Extensive experiments on a number of datasets show the effectiveness of our approach, yielding state-of-the-art results in both quantitative and qualitative benchmarks.*

## 1. Introduction

Enhancing and recovering a high-resolution (HR) image from a low-resolution (LR) counterpart is a theme both of science fiction movies and of the scientific literature. In the latter, it is known as single image super-resolution (SISR), a topic that has enjoyed much attention and progress in recent years. The problem is inherently ill-posed as no unique solution exists: when downsampled, a large number of different HR images can give rise to the same LR image. For high magnification ratios, this one-to-many mapping problem becomes worse, rendering SISR a highly intricate problem. Despite considerable progress in both reconstruction accuracy and speed of SISR, current state-of-the-art methods are still far from image enhancers like the one operated by Harrison Ford alias Rick Deckard in the iconic Blade Runner movie from 1982. A crucial problem is the loss of high-frequency information for large downsampling factors rendering textured regions in super-resolved images blurry, overly smooth, and unnatural in appearance (*c.f.* Fig. 1, left, the new state of the art by PSNR, ENet-E).

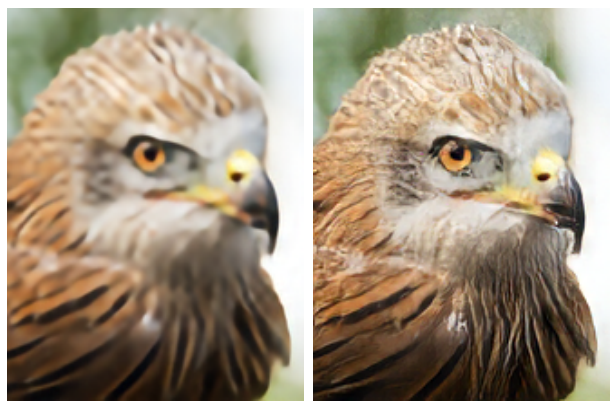

| State of the art by PSNR | Our result |

Figure 1. Comparing the new state of the art by PSNR (ENet-E) with the sharper, perceptually more plausible result produced by ENet-PAT at 4x super-resolution on an image from ImageNet.

The reason for this behavior is rooted in the choice of the objective function that current state-of-the-art methods employ: most systems minimize the pixel-wise mean squared error (MSE) between the HR ground truth image and its reconstruction from the LR observation, which has however been shown to correlate poorly with human perception of image quality [28, 54]. While easy to minimize, the optimal MSE estimator returns the mean of many possible solutions which makes SISR results look unnatural and implausible (*c.f.* Fig. 2). This regression-to-the-mean problem in the context of super-resolution is a well-known fact, however, modeling the high-dimensional multi-modal distribution of natural images remains a challenging problem.

In this work we pursue a different strategy to improve the perceptual quality of SISR results. Using a fully convolutional neural network architecture, we propose a novel modification of recent texture synthesis networks in combination with adversarial training and perceptual losses to produce realistic textures at large magnification ratios. The method works on all RGB channels simultaneously and produces sharp results for natural images at a competitive speed. Trained with suitable combinations of losses, we reach state-of-the-art results both in terms of PSNR and using perceptual metrics.

## 2. Related work

The task of SISR has been studied for decades [23]. Early interpolation methods such as bicubic and Lanczos [11] are based on sampling theory but often produce blurry results with aliasing artifacts in natural images. A large number of high-performing algorithms have since been proposed [35], see also the recent surveys by Nasrollahi and Moeslund [37] and Yang *et al*. [57].

In recent years, popular approaches include exemplar-based models that either exploit recurrent patches of different scales within a single image [13, 17, 22, 56] or learn mappings between low and high resolution pairs of image patches in external databases [3, 5, 14, 27, 51, 58, 63]. They further include dictionary-based methods [33, 40, 52, 59, 61, 64] that learn a sparse representation of image patches as a combination of dictionary atoms, as well as neural network-based approaches [4, 8, 9, 24, 25, 26, 47, 48, 62] which apply convolutional neural networks (CNNs) to the task of SISR. Some approaches are specifically designed for fast inference times [40, 42, 47]. Thus far, realistic textures in the context of high-magnification SISR have only been achieved by user-guided methods [19, 50].

More specifically, Dong *et al*. [8] apply shallow networks to the task of SISR by training a CNN via backpropagration to learn a mapping from the bicubic interpolation of the LR input to a high-resolution image. Later works successfully apply deeper networks and the current state of the art in SISR measured by PSNR is based on deep CNNs [25, 26].

As these models are trained through MSE minimization, the results tend to be blurry and lack high-frequency textures due to the afore-mentioned regression-to-the-mean problem. Alternative perceptual losses have been proposed for CNNs [10, 24] where the idea is to shift the loss from the image-space to a higher-level feature space of an object recognition system like VGG [49], resulting in sharper results despite lower PSNR values.

CNNs have also been found useful for the task of texture synthesis [15] and style transfer [16, 24, 53], however these methods are constrained to the setting of a single network learning to produce only a single texture and have so far not been applied to SISR. Adversarial networks [18] have recently been shown to produce sharp results in a number of image generation tasks [7, 39, 41, 66] but have so far only been applied in the context of super-resolution in a highly constrained setting for the task of face hallucination [62].

Concurrently and independently to our research, in an unpublished work, Ledig *et al*. [29] developed an approach that is similar to ours: inspired by Johnson *et al*. [24], they train feed-forward CNNs using a perceptual loss in conjunction with an adversarial network. However, in contrast to our work, they do not explicitly encourage local matching of texture statistics which we found to be an effective means to produce more realistic textures and to further reduce visually implausible artifacts without the need for additional regularization techniques.

## 3. Single image super-resolution

A high resolution image $I_{\text{HR}} \in [0,1]^{\alpha w \times \alpha h \times c}$ is downsampled to a low resolution image

$$I_{\text{LR}} = \text{d}_\alpha(I_{\text{HR}}) \in [0,1]^{w \times h \times c} \quad (1)$$

using some downsampling operator

$$\text{d}_\alpha : [0,1]^{\alpha w \times \alpha h \times c} \to [0,1]^{w \times h \times c} \quad (2)$$

for a fixed scaling factor $\alpha > 1$, image width $w$, height $h$ and color channels $c$. The task of SISR is to provide an approximate inverse $\text{f} \approx \text{d}^{-1}$ estimating $I_{\text{HR}}$ from $I_{\text{LR}}$:

$$\text{f}(I_{\text{LR}}) = I_{\text{est}} \approx I_{\text{HR}}. \quad (3)$$

This problem is highly ill-posed as the downsampling operation d is non-injective and there exists a very large number of possible images $I_{\text{est}}$ for which $\text{d}(I_{\text{est}}) = I_{\text{LR}}$ holds.

Recent learning approaches aim to approximate f via multi-layered neural networks by minimizing the Euclidean loss $\|I_{\text{est}} - I_{\text{HR}}\|_2^2$ between the current estimate and the ground truth image. While these models reach excellent results as measured by PSNR, the resulting images tend to look blurry and lack high frequency textures present in the original images. This is a direct effect of the high ambiguity in SISR: since downsampling removes high frequency information from the input image, no method can hope to reproduce all fine details with pixel-wise accuracy. Therefore, even state-of-the-art models learn to produce the mean of all possible textures in those regions in order to minimize the Euclidean loss for the output image.

To illustrate this effect, we designed a simple toy example in Fig. 2, where all high frequency information is lost by downsampling. The optimal solution with respect to the Euclidean loss is simply the average of all possible images while more advanced loss functions lead to more realistic, albeit not pixel-perfect reproductions.

| Output size | Layer |
|---|---|
| $w \times h \times c$ | Input $I_{\text{LR}}$ |
| $w \times h \times 64$ | Conv, ReLU |
| | Residual: Conv, ReLU, Conv |
| | ... |
| $2w \times 2h \times 64$ | 2x nearest neighbor upsampling |
| | Conv, ReLU |
| $4w \times 4h \times 64$ | 2x nearest neighbor upsampling |
| | Conv, ReLU |
| | Conv, ReLU |
| | Conv |
| $4w \times 4h \times c$ | Residual image $I_{\text{res}}$ |
| | Output $I_{\text{est}} = I_{\text{bicubic}} + I_{\text{res}}$ |

Table 1. Our generative fully convolutional network architecture for 4x super-resolution which only learns the residual between the bicubic interpolation of the input and the ground truth. We use 3×3 convolution kernels, 10 residual blocks and RGB images ($c = 3$).

## 4. Method

### 4.1. Architecture

Our network architecture in Table 1 is inspired by Long *et al.* [32] and Johnson *et al.* [24] since feed-forward fully convolutional neural networks exhibit a number of useful properties for the task of SISR. The exclusive use of convolutional layers enables training of a single model for an input image of arbitrary size at a given scaling factor $\alpha$ while the feed-forward architecture results in an efficient model at inference time since the LR image only needs to be passed through the network once to get the result. The exclusive use of 3×3 filters is inspired by the VGG architecture [49] and allows for deeper models at a low number of parameters in the network.

As the LR input is smaller than the output image, it needs to be upsampled at some point to produce a high-resolution image estimate. It may seem natural to simply feed the bicubic interpolation of the LR image into the network [8]. However, this introduces redundancies to the input image and leads to a higher computational cost. For convolutional neural networks, Long *et al.* [32] use convolution transpose layers[1] which upsample the feature activations inside the network. This circumvents the nuisance of having to feed a large image with added redundancies into the CNN and allows most computation to be done in the LR image space, resulting in a smaller network and larger receptive fields of the filters relative to the output image.

However, convolution transpose layers have been reported to produce checkerboard artifacts in the output, ne-cessitating an additional regularization term in the output such as total variation [43]. Odena *et al.* [38] replace the convolution transpose layers with nearest-neighbor upsampling of the feature activations in the network followed by a single convolution layer. In our network architecture, this approach still produces checkerboard-artifacts for some specific loss functions, however we found that it obviates the need for an additional regularization term in our more complex models. To further reduce artifacts, we add a convolution layer after all upsampling blocks in the HR image space as this helps to avoid regular patterns in the output.

---
[1]Long *et al.* [32] introduce them as *deconvolution* layers which may be misleading since no actual deconvolution is performed. Other names for convolution transpose layers include *upconvolution*, *fractionally strided convolution* or simply *backwards convolution*.

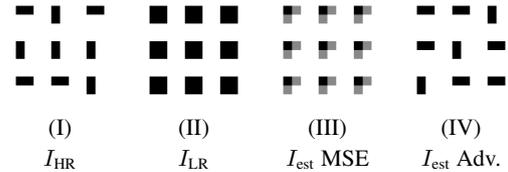

| (I) | (II) | (III) | (IV) |
|---|---|---|---|
| $I_{\text{HR}}$ | $I_{\text{LR}}$ | $I_{\text{est}}$ MSE | $I_{\text{est}}$ Adv. |

Figure 2. Toy example to illustrate the effect of the Euclidean loss and how maximizing the PSNR does not lead to realistic results. (I) The HR images consist of randomly placed vertical and horizontal bars of 1×2 pixels. (II) In $I_{\text{LR}}$, the original orientations cannot be distinguished anymore since both types of bars turn into a single pixel. (III) A model trained to minimize the Euclidean loss produces the mean of all possible solutions since this yields the lowest MSE but the result looks clearly different from the original images $I_{\text{HR}}$. (IV) Training a model with an adversarial loss ideally results in a sharp image that is impossible to distinguish from the original HR images, although it does not match $I_{\text{HR}}$ exactly since the model cannot know the orientation of each bar. Intriguingly, this result has a lower PSNR than the blurry MSE sample.

Training deep networks, we found residual blocks [20] to be beneficial for faster convergence compared to stacked convolution layers. A similarly motivated idea proposed by Kim *et al.* [25] is to learn only the residual image by adding the bicubic interpolation of the input to the model's output, so that it does not need to learn the identity function for $I_{\text{LR}}$. While the residual blocks that make up a main part of our network already only add residual information, we found that applying this idea helps stabilize training and reduce color shifts in the output during training.

### 4.2. Training and loss functions

In this section, we introduce the loss terms used to train our network. Various combinations of these losses and their effects on the results are discussed in Sec. 5.1.

#### 4.2.1 Pixel-wise loss in the image-space

As a baseline, we train our model with the pixel-wise MSE

$$\mathcal{L}_E = ||I_{\text{est}} - I_{\text{HR}}||_2^2, \quad (4)$$

where

$$||I||_2^2 = \frac{1}{whc} \sum_{w,h,c} (I_{w,h,c})^2. \quad (5)$$

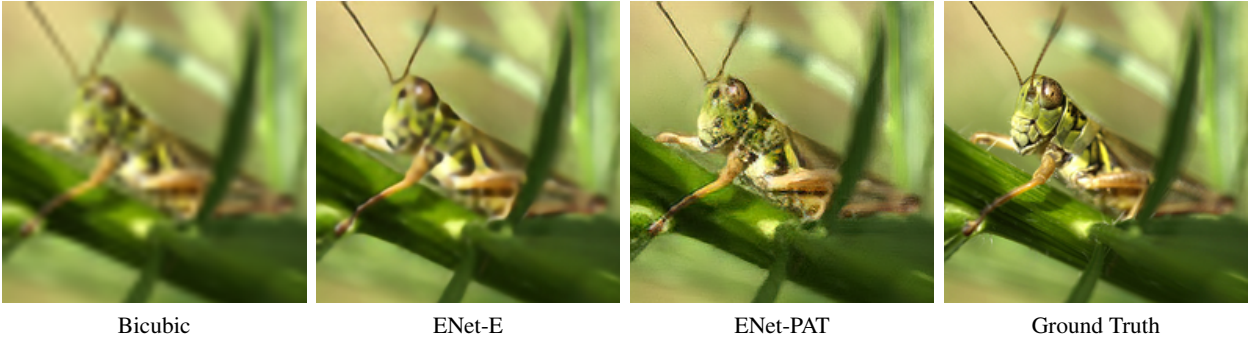

| Bicubic | ENet-E | ENet-PAT | Ground Truth |

Figure 3. Our results on an image from ImageNet for 4x super-resolution. Despite reaching state-of-the-art results by PSNR, ENet-E produces an unnatural and blurry image while ENet-PAT reproduces faithful high-frequency information, resulting in a photorealistic image, at first glance almost indistinguishable from the ground truth image.

#### 4.2.2 Perceptual loss in feature space

Dosovitskiy and Brox [10] as well as Johnson *et al.* [24] propose a *perceptual similarity measure*. Rather than computing distances in image space, both $I_{est}$ and $I_{HR}$ are first mapped into a feature space by a differentiable function $\phi$ before computing their distance.

$$\mathcal{L}_P = ||\phi(I_{est}) - \phi(I_{HR})||_2^2 \qquad (6)$$

This allows the model to generate outputs that may not match the ground truth image with pixel-wise accuracy but instead encourages the network to produce images that have similar feature representations.

For the feature map $\phi$, we use a pre-trained implementation of the popular VGG-19 network [1, 49]. It consists of stacked convolutions coupled with pooling layers to gradually decrease the spatial dimension of the image and to extract higher-level features in higher layers. To capture both low-level and high-level features, we use a combination of the second and fifth pooling layers and compute the MSE on their feature activations.

#### 4.2.3 Texture matching loss

Gatys *et al.* [15, 16] demonstrate how convolutional neural networks can be used to create high quality textures. Given a target texture image, the output image is generated iteratively by matching statistics extracted from a pre-trained network to the target texture. As statistics, correlations between the feature activations $\phi(I) \in \mathbb{R}^{n \times m}$ at a given VGG layer with $n$ features of length $m$ are used:

$$\mathcal{L}_T = ||G(\phi(I_{est})) - G(\phi(I_{HR}))||_2^2, \qquad (7)$$

with Gram matrix $G(F) = FF^T \in \mathbb{R}^{n \times n}$. As it is based on iterative optimization, this method is slow and only works if a target texture is provided at test time. Subsequent works train a feed-forward network that is able to synthesize a global texture (*e.g.*, a given painting style) onto other images [24, 53], however a single network again only produces a single texture, and textures in all input images are replaced by the single style that the network has been trained for.

We propose using the style transfer loss for SISR: Instead of supplying our network with matching high-resolution textures during inference, we compute the texture loss $\mathcal{L}_T$ patch-wise during training to enforce locally similar textures between $I_{est}$ and $I_{HR}$. The network therefore learns to produce images that have the same local textures as the high-resolution images during training. While the task of generating arbitrary textures is more demanding than single-texture synthesis, the LR image and high-level contextual cues give our network more information to work with, enabling it to generate varying high resolution textures. Empirically, we found a patch size of 16×16 pixels to result in the best balance between faithful texture generation and the overall perceptual quality of the images. For results with different patch sizes and further details on the implementation, we refer the reader to the supplementary.

#### 4.2.4 Adversarial training

Adversarial training [18] is a recent technique that has proven to be a useful mechanism to produce realistically looking images. In the original setting, a generative network $G$ is trained to learn a mapping from random vectors $z$ to a data space of images $x$ that is determined by the selected training dataset. Simultaneously, a discriminative network $D$ is trained to distinguish between real images $x$ from the dataset and generated samples $G(z)$. This approach leads to a minimax game in which the generator is trained to minimize

$$\mathcal{L}_A = -\log(D(G(z))) \qquad (8)$$

while the discriminator minimizes

$$\mathcal{L}_D = -\log(D(x)) - \log(1 - D(G(z))). \qquad (9)$$

In the SISR setting, $G$ is our generative network as shown in Fig. 1, *i.e.*, the input to $G$ is now an LR image $I_{LR}$ instead of a noise vector $z$ and its desired output is a suitable realistic high-resolution image $I_{est}$.

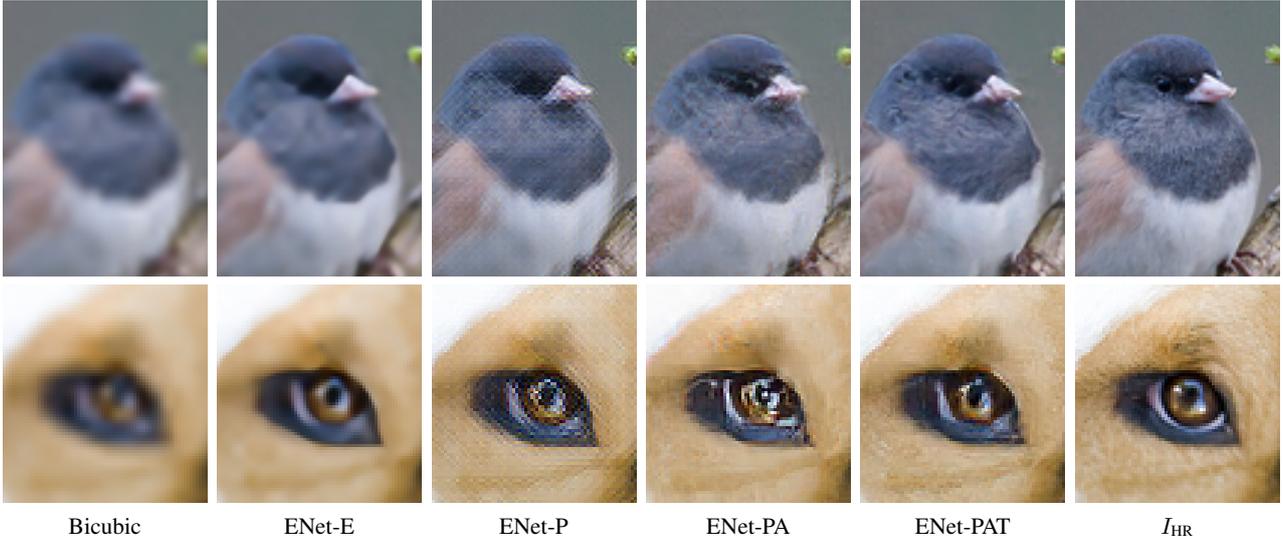

Figure 4. Comparing the results of our model trained with different losses at 4x super-resolution on images from ImageNet. ENet-P's result looks slightly sharper than ENet-E's, but it also produces unpleasing checkerboard artifacts. ENet-PA produces images that are significantly sharper but contain unnatural textures while we found that ENet-PAT generates more realistic textures, resulting in photorealistic images close to the original HR images. Results with further combinations of losses and different parameters are shown in the supplementary.

Following common practice [41], we apply leaky ReLU activations [34] and use strided convolutions to gradually decrease the spatial dimensions of the image in the discriminative network as we found deeper architectures to result in images of higher quality. Perhaps surprisingly, we found dropout not to be effective at preventing the discriminator from overpowering the generator. Instead, the following learning strategy yields better results and a more stable training: we keep track of the average performance of the discriminator on true and generated images within the previous training batch and only train the discriminator in the subsequent step if its performance on either of those two samples is below a threshold. The full architecture and further details are specified in the supplement.

## 5. Evaluation

In Sec. 5.1, we investigate the performance of our architecture trained with different combinations of the previously introduced loss functions. After identifying the best performing models, Sec. 5.2 gives a comprehensive qualitative and quantitative evaluation of our approach. Additional experiments, comparisons and results at various scaling factors are given in the supplementary.

### 5.1. Effect of different losses

We compare the performance of our network trained with the combinations of loss functions listed in Tab. 2. The results are shown in Fig. 4 and Tab. 3 while more results on Enet-EA, ENet-EAT and ENet-PAT trained with different parameters are given in the supplementary.

Using the perceptual loss in ENet-P yields slightly sharper results than ENet-E but it produces artifacts without adding new details in textured areas. Even though the perceptual loss is invariant under perceptually similar transformations, the network is given no incentive to produce realistic textures when trained with the perceptual loss alone.

| Network | Loss | Description |
| --- | --- | --- |
| **ENet-E** | $\mathcal{L}_E$ | Baseline with MSE |
| **ENet-P** | $\mathcal{L}_P$ | Perceptual loss |
| **ENet-EA** | $\mathcal{L}_E + \mathcal{L}_A$ | ENet-E + adversarial |
| **ENet-PA** | $\mathcal{L}_P + \mathcal{L}_A$ | ENet-P + adversarial |
| **ENet-EAT** | $\mathcal{L}_E + \mathcal{L}_A + \mathcal{L}_T$ | ENet-EA + texture loss |
| **ENet-PAT** | $\mathcal{L}_P + \mathcal{L}_A + \mathcal{L}_T$ | ENet-PA + texture loss |

Table 2. The same network trained with varying loss functions.

ENet-PA produces greatly sharper images by adding high frequency details to the output. However, the network sometimes produces unpleasing high-frequency noise to smooth regions and it seems to add high frequencies at random edges resulting in halos and sharpening artifacts in some cases. The texture loss helps ENet-PAT create locally meaningful textures and greatly reduces the artifacts. For some images, the results are almost indistinguishable from the ground truth even at a high magnification ratio of 4.

Unsurprisingly, ENet-E yields the highest PSNR as it is optimized specifically for that measure. Although ENet-PAT produces perceptually more realistic images, the PSNR is much lower as the reconstructions are not pixel-accurate. As shown in the supplementary, SSIM and IFC [46] which have been found to correlate better with human perception [57] also do not capture the perceptual quality of the results, so we provide alternative quantitative evaluations that agree better with human perception in Sec. 5.2.2 and 5.2.3.

| Dataset | Bicubic | ENet-E | ENet-P | ENet-EA | ENet-PA | ENet-EAT | ENet-PAT |
|---------|---------|--------|--------|---------|---------|----------|----------|
| Set5    | 28.42   | **31.74** | 28.28 | 28.15 | 27.20 | 29.26 | 28.56 |
| Set14   | 26.00   | **28.42** | 25.64 | 25.94 | 24.93 | 26.53 | 25.77 |
| BSD100  | 25.96   | **27.50** | 24.73 | 25.71 | 24.19 | 25.97 | 24.93 |
| Urban100| 23.14   | **25.66** | 23.75 | 23.56 | 22.51 | 24.16 | 23.54 |

Table 3. PSNR for our architecture trained with different combinations of losses at 4x super resolution. ENet-E yields the highest PSNR values since it is trained towards minimizing the per-pixel distance to the ground truth. The models trained with the perceptual loss all yield lower PSNRs as it allows for deviations in pixel intensities from the ground truth. It is those outliers that significantly lower the PSNR scores. The texture loss increases the PSNR values by reducing the artifacts from the adversarial loss term. Best results shown in bold.

## 5.2. Comparison with other approaches

Figure 5 gives an overview of different approaches including the current state of the art by PSNR [25, 26] on the zebra image from Set14 which is particularly well-suited for a visual comparison since it contains both smooth and sharp edges, textured regions as well as repeating patterns. Previous methods have gradually improved on edge reconstruction, but even the state-of-the-art model DRCN suffers from blur in regions where the LR image doesn't provide any high frequency information. While ENet-E reproduces slightly sharper edges, the results exhibit the same characteristics as previous approaches. The perceptual loss from Johnson et al. [24] produces only a slightly sharper image than ENet-E. On the other hand, ENet-PAT is the only model that produces significantly sharper images with realistic textures. Comparisons with further works including Johnson et al. [24], Bruna et al. [4] and Romano et al. [42] are shown in Fig. 6 and in the supplementary.

### 5.2.1 Quantitative results by PSNR

Table 4 summarizes the PSNR values of our approach in comparison to other approaches including the previous state of the art on various popular SISR benchmarks. ENet-E achieves state-of-the-art results on all datasets.

### 5.2.2 Object recognition performance

It is known that super-resolution algorithms can be used as a preprocessing step to improve the performance of other image-related tasks such as face recognition [12]. We propose to use the performance of state-of-the-art object recognition models as a metric to evaluate image reconstruction algorithms, especially for models whose performance is not captured well by PSNR, SSIM or IFC.

For evaluation, any pre-trained object recognition model $M$ and labeled set of images may be used. The image restoration models to be evaluated are applied on a degraded version of the dataset and the reconstructed images are fed into $M$. The hypothesis is that the performance of powerful object recognition models shows a meaningful correlation with the human perception of image quality that may complement pixel-based benchmarks such as PSNR.

Similar indirect metrics have been applied in previous works, *e.g.*, optical character recognition performance has been utilized to compare the quality of text deblurring algorithms [21, 55] and face-detection performance has been used for the evaluation of super-resolution algorithms [30]. The performance of object recognition models has been used for the indirect evaluation of image colorization [65], where black and white images were colorized to improve object detection rates. Namboodiri *et al.* [36] apply a metric similar to ours to evaluate SISR algorithms and found it to be a better metric than PSNR or SSIM for evaluating the perceptual quality of super-resolved images.

For our comparison, we use ResNet-50 [6, 20] as this class of models has achieved state-of-the-art performance by winning the 2015 Large Scale Visual Recognition Challenge (ILSVRC) [44]. For the evaluation, we use the first 1000 images in the ILSVRC 2016 CLS-LOC validation dataset[2] where each image has exactly one out of 1000 labels. The original images are scaled to 224×224 for the baseline and downsampled to 56×56 for a scaling factor of 4. We report the mean top-1 and top-5 errors as well as the mean confidence that ResNet reports on correct classifications. The results are shown in Tab. 5. In our comparison, some of the results roughly coincide with the PSNR scores, with bicubic interpolation resulting in the worst performance followed by DRCN [26] and PSyCo [40] which yield visually comparable images and hence similar scores as our ENet-E network. However, our models ENet-EA, ENet-PA and ENet-PAT produce images of higher perceptual quality which is reflected in higher classification scores despite their low PSNR scores. This indicates that the object recognition benchmark matches human perception better than PSNR does. The high scores of ENet-PAT are not a result of overfitting due to being trained with VGG, since even ENet-EA (which is not trained with VGG) gains higher scores than e.g. ENet-E, which has the highest PSNR but lower scores under this metric.

While we observe that the object recognition performance roughly coincides with the human perception of image quality in this benchmark for super-resolution, we leave a more detailed analysis of this evaluation metric on other image restoration problems to future work.

---

[2]We use the validation dataset since the annotations for the test dataset are not released. However, even a potential bias of the ResNet-model would not invalidate the results, since higher scores only imply that the upscaled images are closer to the originals under the proposed metric.

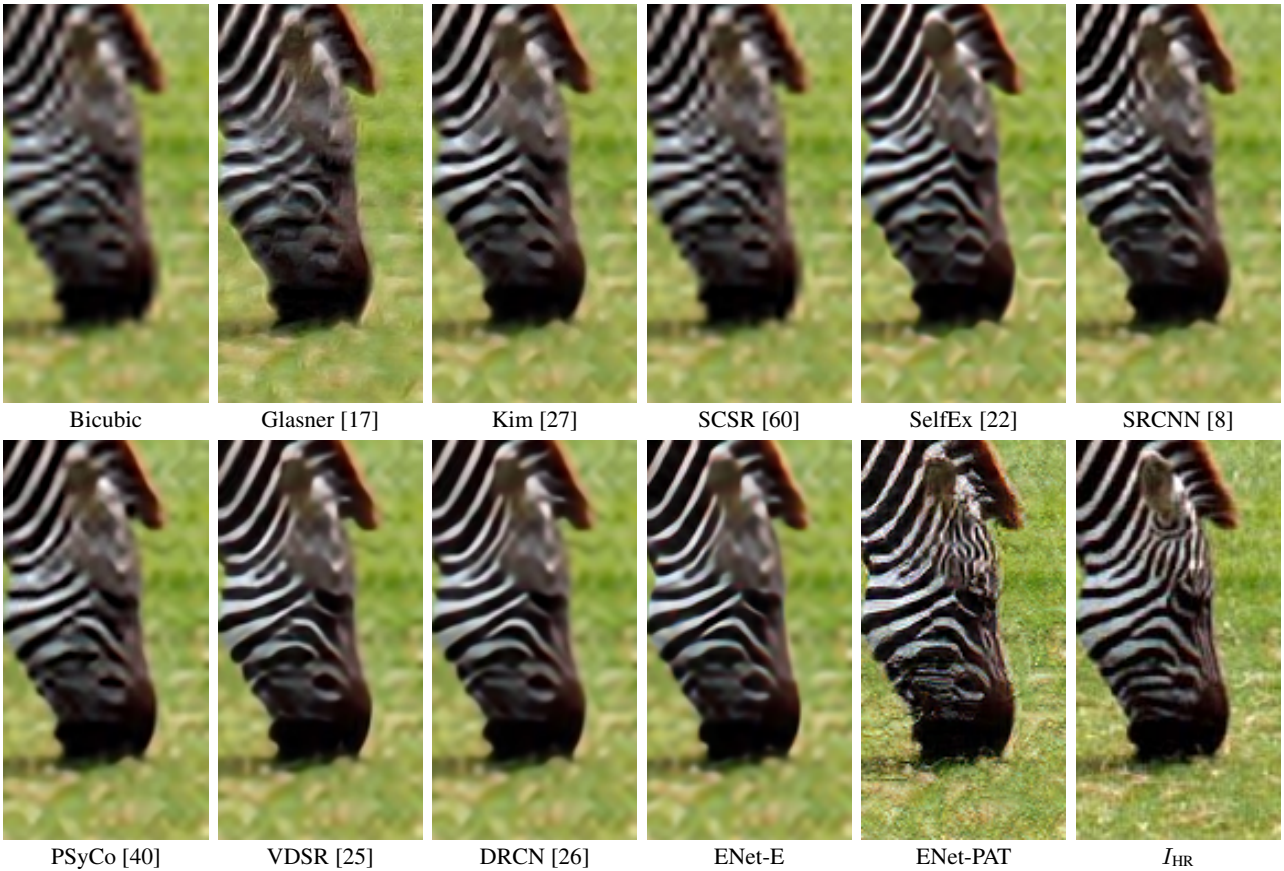

Figure 5. A comparison of previous methods with our results at 4x super-resolution on an image from Set14. Previous methods have continuously improved upon the restoration of sharper edges yielding higher PSNR's, a trend that ENet-E continues with slightly sharper edges and finer details (*e.g.*, area below the eye). With our texture-synthesizing approach, ENet-PAT is the only method that yields sharp lines and reproduces textures, resulting in the most realistic looking image. Furthermore, ENet-PAT produces high-frequency patterns missing completely in the LR image, *e.g.*, lines on the zebra's forehead or the grass texture, showing that the model is capable of detecting and generating patterns that lead to a realistic image.

| $\alpha = 4$ Dataset | Bicubic Baseline | RFL [45] | A+ [51] | SelfEx [22] | SRCNN [8] | PSyCo [40] | ESPCN [47] | DRCN [26] | VDSR [25] | ENet-E ours |
|---|---|---|---|---|---|---|---|---|---|---|
| Set5 | 28.42 | 30.14 | 30.28 | 30.31 | 30.48 | 30.62 | 30.90 | 31.53 | 31.35 | **31.74** |
| Set14 | 26.00 | 27.24 | 27.32 | 27.40 | 27.49 | 27.57 | 27.73 | 28.02 | 28.01 | **28.42** |
| BSD100 | 25.96 | 26.75 | 26.82 | 26.84 | 26.90 | 26.98 | – | 27.23 | 27.29 | **27.50** |
| Urban100 | 23.14 | 24.19 | 24.32 | 24.79 | 24.52 | 24.62 | – | 25.14 | 25.18 | **25.66** |

Table 4. PSNR for different methods at 4x super-resolution. ENet-E achieves state-of-the-art results on all datasets. Best performance shown in bold. Further results as well as SSIM and IFC scores on varying scaling factors are given in the supplementary.

### 5.2.3 Evaluation of perceptual quality

To further validate the perceptual quality of our results, we conducted a user study on the ImageNet dataset from the previous section. As a representative for models that minimize the Euclidean loss, we compare ENet-E as the new state of the art in PSNR performance with the images generated by ENet-PAT which have a PSNR comparable to images upsampled with bicubic interpolation. The subjects were shown the ground truth image along with the super-resolution results of both ENet-E and ENet-PAT at 4x super-resolution side-by-side, and were asked to select the image that looks more similar to the ground truth. In 49 survey responses for a total of 843 votes, subjects selected the image produced by ENet-PAT 91.0% of the time, underlining the perceptual quality of our results. For a screenshot of the survey and an analysis on the images where the blurry result by ENet-E was prefered, we refer the reader to the supplementary material.

| Evaluation | Bicubic | DRCN [26] | PSyCo [40] | ENet-E | ENet-EA | ENet-PA | ENet-PAT | Baseline |
|---|---|---|---|---|---|---|---|---|
| Top-1 error | 0.506 | 0.477 | 0.454 | 0.449 | 0.407 | 0.429 | **0.399** | 0.260 |
| Top-5 error | 0.266 | 0.242 | 0.224 | 0.214 | 0.185 | 0.199 | **0.171** | 0.072 |
| Confidence | 0.754 | 0.727 | 0.728 | 0.754 | 0.760 | 0.783 | **0.797** | 0.882 |

Table 5. ResNet object recognition performance and reported confidence on pictures from the ImageNet dataset downsampled to 56×56 before being upscaled by a factor of 4 using different algorithms. The baseline shows ResNet's performance on the original 224×224 sized images. Compared to PSNR, the scores correlate better with the human perception of image quality: ENet-E achieves only slightly higher scores than DRCN or PSyCo since all these models minimize pixel-wise MSE. On the other hand, ENet-PAT achieves higher scores as it produces sharper images and more realistic textures. The good results of ENet-EA which is trained without VGG indicate that the high scores of ENet-PAT are not solely due to being trained with VGG, but likely a result of sharper images. Best results shown in bold.

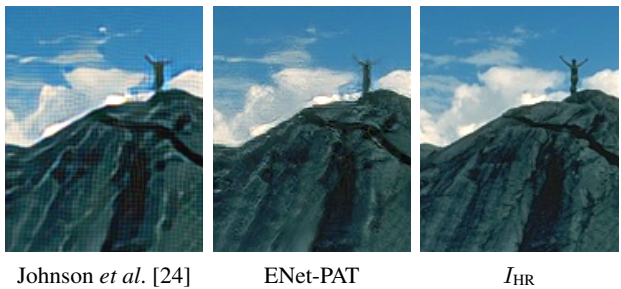

Johnson *et al*. [24]  ENet-PAT  $I_{\text{HR}}$

Figure 6. Comparing our model with a result from Johnson *et al*. [24] on an image from BSD100 at 4x super-resolution. ENet-PAT's result looks more natural and does not contain checkerboard artifacts despite the lack of an additional regularization term.

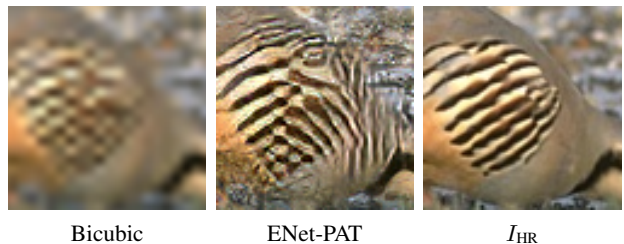

Bicubic  ENet-PAT  $I_{\text{HR}}$

Figure 7. Failure case on an image from BSD100. ENet-PAT has learned to continue high-frequency patterns since they are often lost in $I_{\text{LR}}$ at smaller scales. While that works out extremely well in most cases (*c.f.* zebra's forehead Fig. 5), the model fails in this notable case since $I_{\text{HR}}$ is actually smooth in that region.

### 5.3. Training and inference speed

For training, we use all color images in MSCOCO [31] that have at least 384 pixels on the short side resulting in roughly 200k images. All images are cropped centrally to a square and then downsampled to 256×256 to reduce noise and JPEG artifacts. During training, we fix the size of the input $I_{\text{LR}}$ to 32×32. As the scale of objects in the MSCOCO dataset is too small when downsampled to such a small size, we downsample the 256×256 images by $\alpha$ and then crop these to patches of size 32×32. After training the model for any given scaling factor $\alpha$, the input to the fully convolutional network at test time can be an image of arbitrary dimensions $w \times h$ which is then upscaled to $(\alpha w) \times (\alpha h)$.

We trained all models for a maximum of 24 hours on an Nvidia K40 GPU using TensorFlow [2], though convergence rates depend on the applied combination of loss functions. Although not optimized for efficiency, our network is compact and quite fast at test time. The final trained model is only 3.1MB in size and processes images in 9ms (Set5), 18ms (Set14), 12ms (BSD100) and 59ms (Urban100) on average per image at 4x super-resolution.

## 6. Discussion, limitations and future work

We have proposed an architecture that is capable of producing state-of-the-art results by both quantitative and qualitative measures by training with a Euclidean loss or a novel combination of adversarial training, perceptual losses and a newly proposed texture transfer loss for super-resolution. Once trained, the model interpolates full color images in a single forward-pass at competitive speeds.

As SISR is a heavily ill-posed problem, some limitations remain. While images produced by ENet-PAT look realistic, they do not match the ground truth images on a pixel-wise basis. Furthermore, the adversarial training sometimes produces artifacts in the output which are greatly reduced but not fully eliminated with the addition of the texture loss.

We noted an interesting failure on an image in the BSD100 dataset that is shown in Fig. 7, where the model continues a pattern visible in the LR image onto smooth areas. This is a result of the model learning to hallucinate textures that occur frequently between pairs of LR and HR images such as repeating stripes that fade in the LR image as they increasingly shrink in size.

While the model is already competitive in terms of its runtime, future work may decrease the depth of the network and apply shrinking methods to speed up the model to real-time performance on high-resolution data: adding a term for temporal consistency could then enable the model to be used for video super-resolution. We refer the reader to the supplementary material for more results, further details and additional comparisons. A reference implementation of ENet-PAT can be found on the project website at http://webdav.tue.mpg.de/pixel/enhancenet.

# EnhanceNet: Single Image Super-Resolution Through Automated Texture Synthesis — Supplementary —

Mehdi S. M. Sajjadi    Bernhard Schölkopf    Michael Hirsch

Max Planck Institute for Intelligent Systems
Spemanstr. 34, 72076 Tübingen, Germany

{msajjadi,bs,mhirsch}@tue.mpg.de

## Abstract

*In this supplemental, we present some further details on our models and their training procedure, provide additional insights about the influence of the different loss functions to the super-resolution reconstruction, discuss applications and limitations of our approach and show further results and comparisons with other methods. The sections in the supplementary are numbered to match the corresponding sections in the main paper.*

## 4 Additional details on the method

### 4.2.3 Patch size of texture matching loss

We compute the texture loss $\mathcal{L}_T$ patch-wise to enforce locally similar textures between $I_{\text{est}}$ and $I_{\text{HR}}$. We found a patch size of 16×16 pixels to result in the best balance between faithful texture generation and the overall perceptual quality of the images. Figure 1 shows ENet-PAT when trained using patches of size 4×4 pixels for the texture matching loss (ENet-PAT-4) and when it is calculated on larger patches of 128×128 pixels (ENet-PAT-128). Using smaller patches leads to artifacts in textured regions while calculating the texture matching loss on too large patches during training leads to artifacts throughout the entire image since the network is trained with texture statistics that are averaged over regions of varying textures, leading to unpleasant results.

### 4.2.4 Architecture of the adversarial network

Table 1 shows the architecture of our discriminative adversarial network used for the loss term $\mathcal{L}_A$. We follow common design patterns [13] and exclusively use convolutional layers with filters of size 3×3 pixels with varying stride lengths to reduce the spatial dimension of the input down to a size of 4×4 pixels where we append two fully connected layers along with a sigmoid activation at the output to produce a classification label between 0 and 1.

| Output size | Layer |
| --- | --- |
| 128 × 128 × 3 | Input $I_{\text{est}}$ or $I_{\text{HR}}$ |
| 128 × 128 × 32 | Conv, lReLU |
| 64 × 64 × 32 | Conv stride 2, lReLU |
| 64 × 64 × 64 | Conv, lReLU |
| 32 × 32 × 64 | Conv stride 2, lReLU |
| 32 × 32 × 128 | Conv, lReLU |
| 16 × 16 × 128 | Conv stride 2, lReLU |
| 16 × 16 × 256 | Conv, lReLU |
| 8 × 8 × 256 | Conv stride 2, lReLU |
| 8 × 8 × 512 | Conv, lReLU |
| 4 × 4 × 512 | Conv stride 2, lReLU |
| 8192 | Flatten |
| 1024 | Fc, lReLU |
| 1 | Fc, sigmoid |
| 1 | Estimated label |

Table 1. The network architecture of our adversarial discriminative network at 4x super-resolution. As in the generative network, we exclusively use 3×3 convolution kernels. The network design draws inspiration from VGG [17] but uses leaky ReLU activations [11] and strided convolutions instead of pooling layers [13].

## 5 Further evaluation of results

Our models only learn the residual image between the bicubic upsampled input image and the high resolution output which renders training more stable. Figure 3 displays examples for residual images that our models estimate. ENet-E has learned to significantly increase the sharpness of the image and to remove aliasing effects in the bicubic interpolation (as seen in the aliasing effects in the residual image that cancel out with the aliasing in the bicubic interpolation). ENet-PAT additionally generates fine high-frequency textures in regions that should be textured while leaving smooth areas such as the sky and the red front areas of the house untouched.

### 5.1 Additional combinations of losses

In general, we found training models with the adversarial and texture matching loss in conjunction with the Euclidean loss (in place of the perceptual loss) to be significantly less stable and the perceptual quality of the results oscillated heavily during training, *i.e.*, ENet-EA and ENet-EAT are harder to train than ENet-PA and ENet-PAT. This is because the adversarial and texture losses encourage the synthesis of high frequency information in the results, increasing the Euclidean distance to the ground truth images during training which leads to loss functions that counteract each other. The perceptual loss on the other hand is more tolerant to small-scale deviations due to pooling. The results of ENet-EA and ENet-EAT are shown in Fig. 2. We note that the texture matching loss in ENet-EAT leads to a more stable training than ENet-EA and slightly better results, though worse than ENet-PAT. This means that the texture matching loss not only helps create more realistic textures, but it also stabilizes the adversarial training to an extent.

### 5.2 Comparison with further methods

Figure 5 shows a comparison of our method with Bruna *et al.* [2]. Our model does not suffer from jagged edges and is much sharper.

Figure 6 shows a comparison with RAISR [14] at 2x super-resolution. Since RAISR has been designed for speed rather than state-of-the-art image quality, it reaches a lower performance than previous methods [7, 8, 12] so ENet-E yields visually sharper images even at this low scaling factor. ENet-PAT is the only model to reconstruct sharp details and it is visually much less distinguishable from the ground truth. Despite not being optimized for speed, EnhanceNet is even faster than RAISR at test-time: 9/18ms (EnhanceNet) vs. 17/30ms (RAISR) on average per image at 4x super-resolution on Set5/Set14, though EnhanceNet runs on a GPU while RAISR has been benchmarked on a 6-core CPU.

To demonstrate the performance of our method, we compare the result of ENet-PAT at 4x super-resolution with the current state of the art models at 2x super-resolution in Fig. 4. Although 4x super-resolution is a greatly more demanding task than 2x super-resolution, the results are comparable in quality. Small details that are lost completely in the 4x downsampled image are more accurate in VDSR and DRCN's outputs, but our model produces a plausible image with sharper textures at 4x super-resolution that even outperforms the current state of the art at 2x super-resolution in sharpness, *e.g.*, the area below the eyes is sharper in ENet-PAT's result and looks very similar to the ground truth.

| Model | Loss | Weight | VGG layer |
|---|---|---|---|
| ENet-P | $\mathcal{L}_P$ | $2 \cdot 10^{-1}$ | $pool_2$ |
|  |  | $2 \cdot 10^{-2}$ | $pool_5$ |
| ENet-PA | $\mathcal{L}_P$ | $2 \cdot 10^{-1}$ | $pool_2$ |
|  |  | $2 \cdot 10^{-2}$ | $pool_5$ |
|  | $\mathcal{L}_A$ | 1 | – |
| ENet-PAT | $\mathcal{L}_P$ | $2 \cdot 10^{-1}$ | $pool_2$ |
|  |  | $2 \cdot 10^{-2}$ | $pool_5$ |
|  | $\mathcal{L}_A$ | 2 | – |
|  | $\mathcal{L}_T$ | $3 \cdot 10^{-7}$ | $conv_{1.1}$ |
|  |  | $1 \cdot 10^{-6}$ | $conv_{2.1}$ |
|  |  | $1 \cdot 10^{-6}$ | $conv_{3.1}$ |

Table 2. Weights for the losses used to train our models.

#### 5.2.1 Quantitative results by PSNR, SSIM and IFC

Tables 3, 4 and 5 show quantitative results measured by PSNR, SSIM and IFC [16] for varying scaling factors. None of these metrics is able to correctly capture the perceptual quality of ENet-PAT's results.

#### 5.2.3 Screenshot of the survey

Figure 7 shows a screenshot of the survey that we used to evaluate the perceptual quality of our results. The subjects were shown the target image on the top and were asked to click the image on the bottom that looks more similar to the target image. Each subject was shown up to 30 images.

### 5.3 Implementation details and training

The model has been implemented in TensorFlow r0.10 [1]. For all weights, we apply Xavier initialization [5]. For training, we use the Adam optimizer [9] with an initial learning rate of $10^{-4}$. We found common convolutional layers stacked with ReLU's to yield comparable results, but training converges faster with the residual architecture. All models were trained only once and used for all results throughout the paper and the supplementary, no fine-tuning was done for any specific dataset or image. Nonetheless, we believe that a choice of specialized training datasets for specific types of images can greatly increase the perceptual quality of the produced textures (*c.f.* Sec. 6).

For the perceptual loss $\mathcal{L}_P$ and the texture loss $\mathcal{L}_T$, we normalized feature activations to have a mean of one [4]. For the texture matching loss, we use a combination of the first convolution in each of the first three groups of layers in VGG, similar to Gatys *et al.* [4]. For the weights, we chose the combination that produced the most realistically looking results. The exact values of the weights for the different losses are given in Table 2.

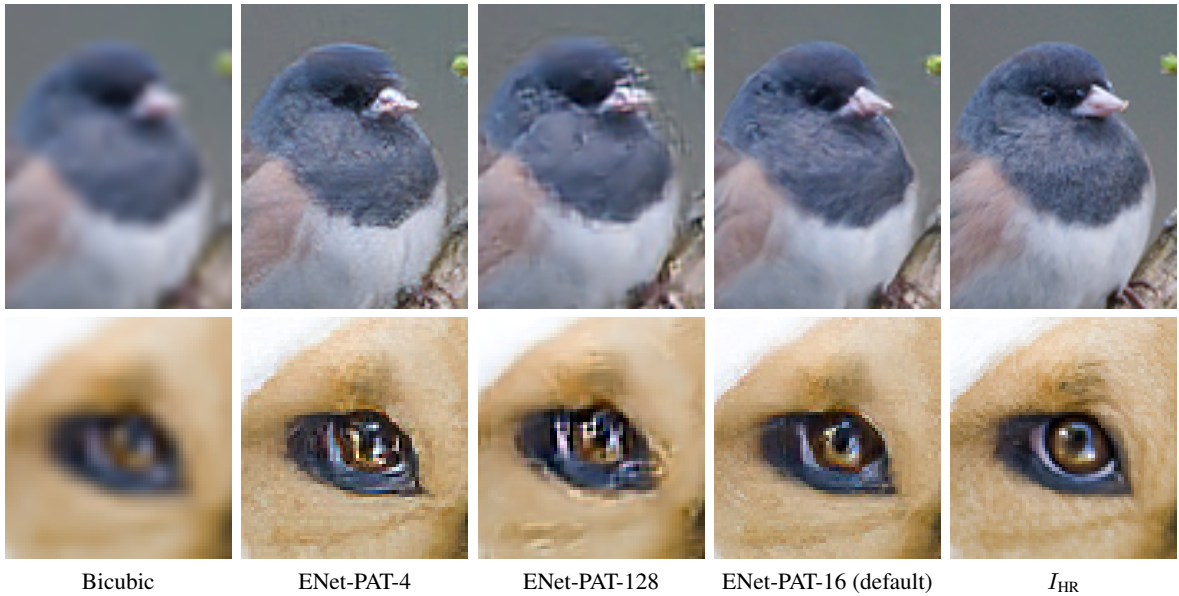

Figure 1. Comparing different patch sizes for the texture matching loss during training for ENet-PAT on images from ImageNet at 4x super-resolution. Computing the texture matching loss on small patches fails to capture textures properly (ENet-PAT-4) while matching textures on the whole image leads to unpleasant results since different texture statistics are averaged (ENet-PAT-128).

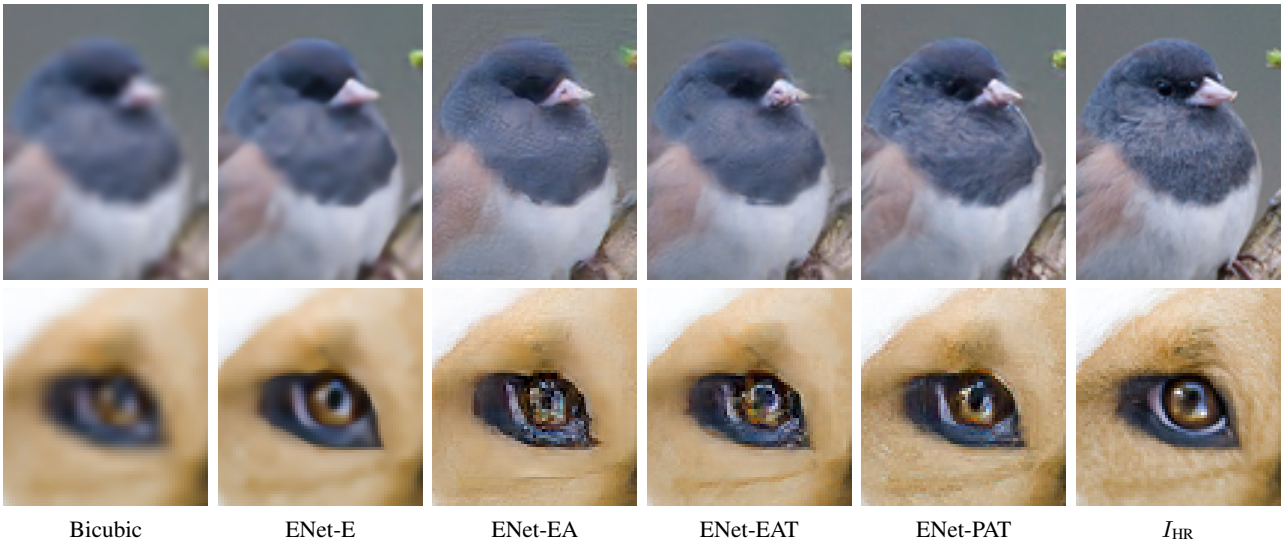

Figure 2. Replacing the perceptual loss in ENet-PA and ENet-PAT with the Euclidean loss results in images with sharp but jagged edges and overly smooth textures (4x super-resolution). Furthermore, these models are significantly harder to train.

## 6 Specialized training datasets

Figure 8 shows an example for an image where the majority of subjects in our survey preferred ENet-E's result over the image produced by ENet-PAT. In general, ENet-PAT trained on MSCOCO struggles to reproduce realistically looking faces at high scaling factors and while the overall image is significantly sharper than the result of ENet-E, the human perception is highly sensitive to small changes in the appearance of human faces which is why many subjects preferred the blurry result of ENet-E in those cases. To demonstrate that this is not a limitation of our model, we train ENet-PAT with identical hyperparameters on the CelebA dataset [10] (ENet-PAT-F) and compare the results with ENet-PAT trained on MSCOCO as before. The results are shown in Fig. 9. When trained on CelebA, ENet-PAT-F has significantly better performance.

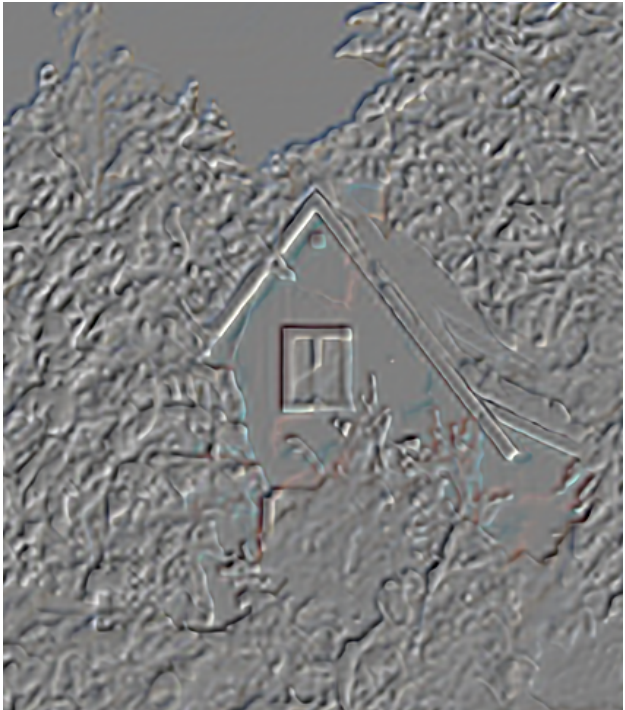
ENet-E residual

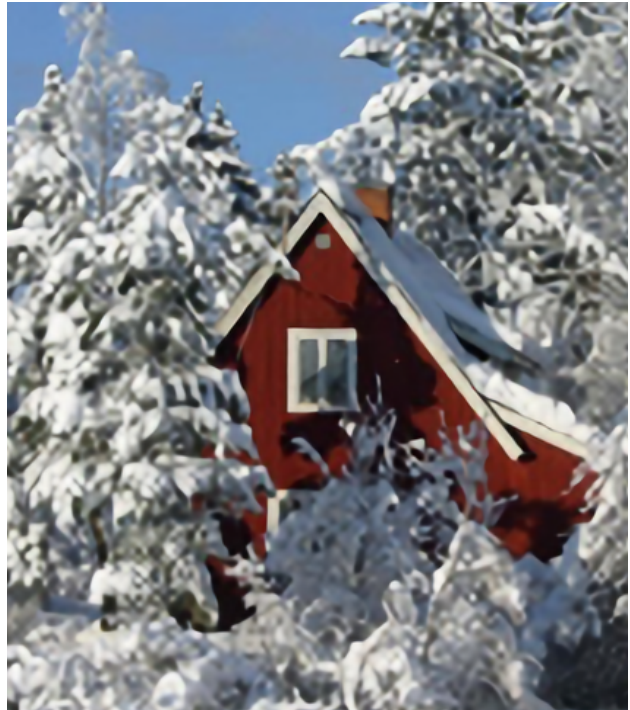
ENet-E result

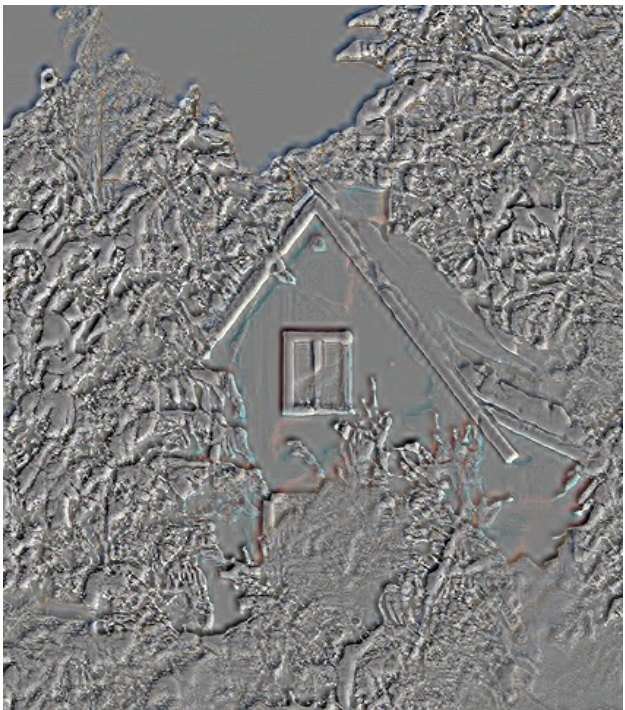
ENet-PAT residual

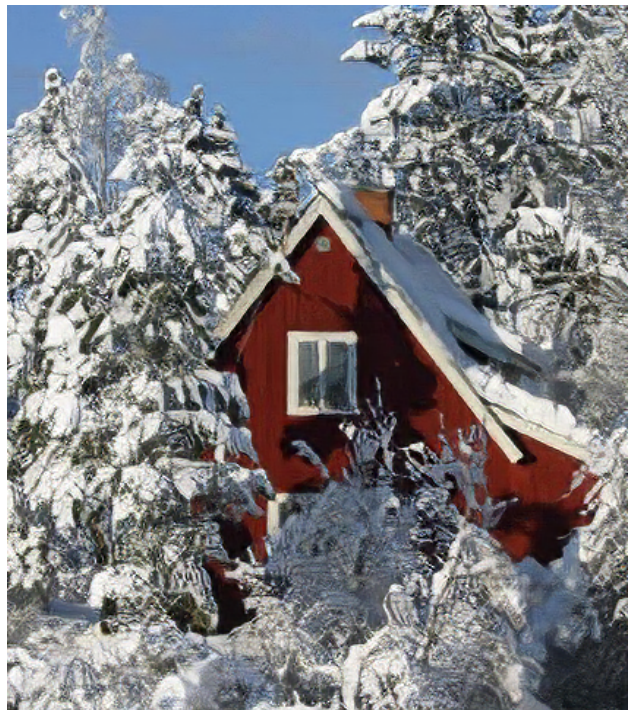
ENet-PAT result

Figure 3. A visualization of the residual image that the network produces at 4x super-resolution. While ENet-E significantly sharpens edges and is able to remove aliasing from the bicubic interpolation, ENet-PAT produces additional textures yielding a sharp, realistic result. Image taken from the SunHays80 dataset [18].

| 2x downsampled input | 2x downsampled input | 4x downsampled input | $I_{\text{HR}}$ |
|---|---|---|---|
| ↓ | ↓ | ↓ | |
| 2x VDSR [7] | 2x DRCN [8] | 4x ENet-PAT | $I_{\text{HR}}$ |

Figure 4. Comparing the previous state of the art by PSNR value at 2x super-resolution (75% of all pixels missing) with our model at 4x super-resolution (93.75% of all pixels missing). The top row shows the input to the models and the bottom row the results. Although our model has significantly less information to work with, it produces a sharper image with realistic textures.

| Scatter [2] | Fine-tuned scatter [2] | VGG [2] | ENet-PAT | $I_{\text{HR}}$ |
|---|---|---|---|---|

Figure 5. Comparing our model with Bruna *et al.* [2] at 4x super-resolution. ENet-PAT produces images with more contrast and sharper edges that are more faithful to the ground truth.

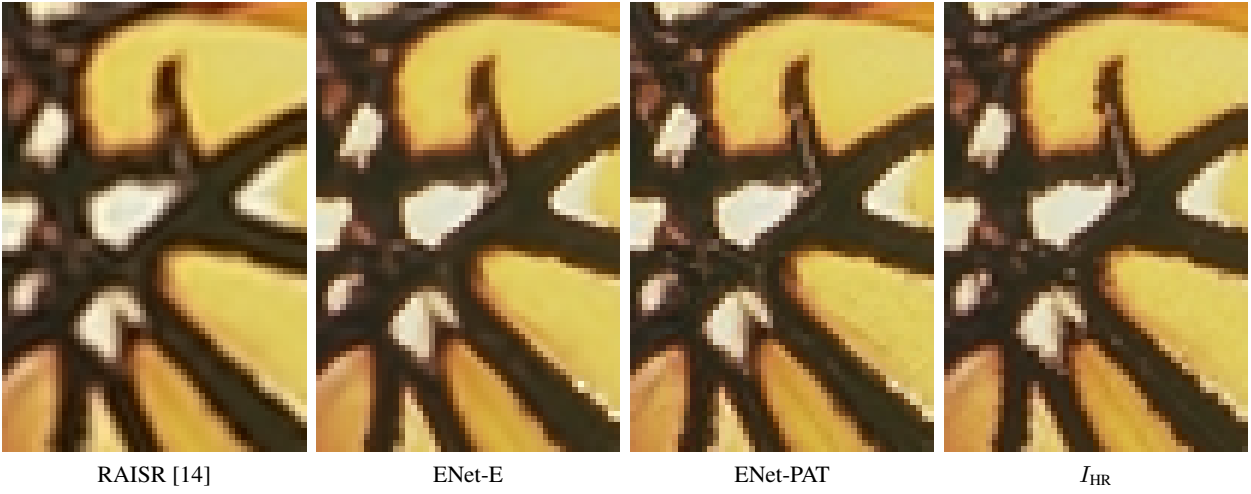

| RAISR [14] | ENet-E | ENet-PAT | $I_{HR}$ |

Figure 6. Comparing our model with Romano *et al*. [14] at 2x super-resolution on the butterfly image of Set5. Despite the low scaling factor, image quality gradually increases between RAISR, ENet-E and ENet-PAT, the last of which is not only sharper but also recreates small details better, *e.g*., the vertical white line in the middle of the picture is fully reconstructed only in ENet-PAT's result.

| $\alpha = 2$ Dataset | Bicubic Baseline | RFL [15] | A+ [19] | SelfEx [6] | SRCNN [3] | PSyCo [12] | DRCN [8] | VDSR [7] | ENet-E ours | Enet-PAT ours |
|---|---|---|---|---|---|---|---|---|---|---|
| Set5 | 33.66 | 36.54 | 30.14 | 36.49 | 36.66 | 36.88 | **37.63** | 37.53 | 37.32 | 33.89 |
| Set14 | 30.24 | 32.26 | 27.24 | 32.22 | 32.42 | 32.55 | 33.04 | 33.03 | **33.25** | 30.45 |
| BSD100 | 29.56 | 31.16 | 26.75 | 31.18 | 31.36 | 31.39 | 31.85 | 31.90 | **31.95** | 28.30 |
| Urban100 | 26.88 | 29.11 | 24.19 | 29.54 | 29.50 | 29.64 | 30.75 | 30.76 | **31.21** | 29.00 |

Table 3. PSNR for different methods at 2x super-resolution. Best performance shown in bold.

| $\alpha = 2$ Dataset | Bicubic Baseline | RFL [15] | A+ [19] | SelfEx [6] | SRCNN [3] | PSyCo [12] | DRCN [8] | VDSR [7] | ENet-E ours | Enet-PAT ours |
|---|---|---|---|---|---|---|---|---|---|---|
| Set5 | 0.9299 | 0.9537 | 0.9544 | 0.9537 | 0.9542 | 0.9559 | **0.9588** | 0.9587 | 0.9581 | 0.9276 |
| Set14 | 0.8688 | 0.9040 | 0.9056 | 0.9034 | 0.9063 | 0.8984 | 0.9118 | 0.9124 | **0.9148** | 0.8617 |
| BSD100 | 0.8431 | 0.8840 | 0.8863 | 0.8855 | 0.8879 | 0.8895 | 0.8942 | 0.8960 | **0.8981** | 0.8729 |
| Urban100 | 0.8403 | 0.8706 | 0.8938 | 0.8947 | 0.8946 | 0.9000 | 0.9133 | 0.9140 | **0.9194** | 0.8303 |

| $\alpha = 4$ Dataset | Bicubic Baseline | RFL [15] | A+ [19] | SelfEx [6] | SRCNN [3] | PSyCo [12] | DRCN [8] | VDSR [7] | ENet-E ours | Enet-PAT ours |
|---|---|---|---|---|---|---|---|---|---|---|
| Set5 | 0.8104 | 0.8548 | 0.8603 | 0.8619 | 0.8628 | 0.8678 | 0.8854 | 0.8838 | **0.8869** | 0.8082 |
| Set14 | 0.7027 | 0.7451 | 0.7491 | 0.7518 | 0.7503 | 0.7525 | 0.8670 | 0.7674 | **0.7774** | 0.6784 |
| BSD100 | 0.6675 | 0.7054 | 0.7087 | 0.7106 | 0.7101 | 0.7159 | 0.7233 | 0.7251 | **0.7326** | 0.6270 |
| Urban100 | 0.6577 | 0.7096 | 0.7183 | 0.7374 | 0.7221 | 0.7317 | 0.7510 | 0.7524 | **0.7703** | 0.6936 |

Table 4. SSIM for different methods at 2x and 4x super-resolution. Similar to PSNR, ENet-PAT also yields low SSIM values despite the perceptual quality of its results. Best performance shown in bold.

| $\alpha = 4$ Dataset | Bicubic Baseline | RFL [15] | A+ [19] | SelfEx [6] | SRCNN [3] | PSyCo [12] | DRCN [8] | VDSR [7] | ENet-E ours | ENet-PAT ours |
|---|---|---|---|---|---|---|---|---|---|---|
| Set5 | 2.329 | 3.191 | 3.248 | 3.166 | 2.991 | 3.379 | **3.554** | 3.553 | 3.413 | 2.643 |
| Set14 | 2.237 | 2.919 | 2.751 | 2.893 | 2.751 | 3.055 | 3.112 | **3.122** | 3.093 | 2.281 |
| Urban100 | 2.361 | 3.110 | 3.208 | 3.314 | 2.963 | 3.351 | 3.461 | 3.459 | **3.508** | 2.635 |

Table 5. IFC for different methods at 4x super-resolution. Best performance shown in bold. The IFC scores roughly follow PSNR and do not capture the perceptual quality of ENet-PAT's results.

# Image Quality Assessment

**30 images to go!**

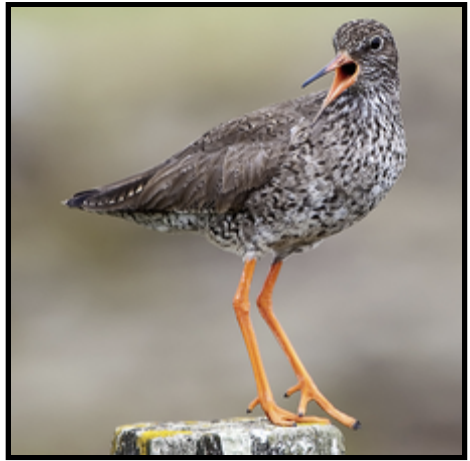

Target Image

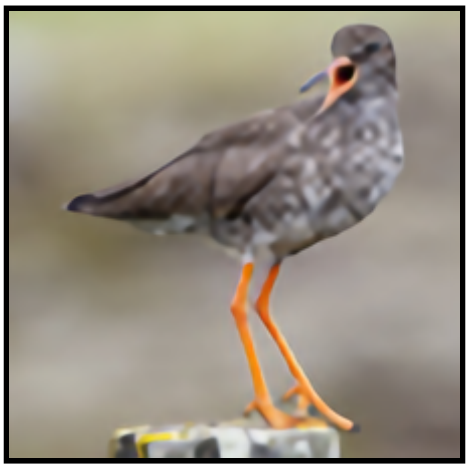
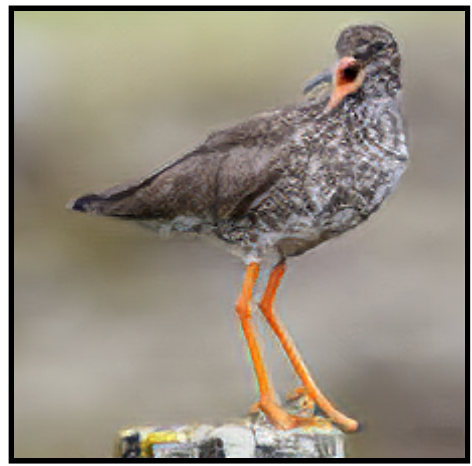

Click the image that looks more similar to the target image above.

Figure 7. Example screenshot of our survey for perceptual image quality. Subjects were shown a target image above and were asked to select the image on the bottom that looks more similar to the target image. In 49 survey responses for a total of 843 votes, subjects selected the image produced by ENet-PAT 91.0%, underlining its higher perceptual quality compared to the state of the art by PSNR, ENet-E.

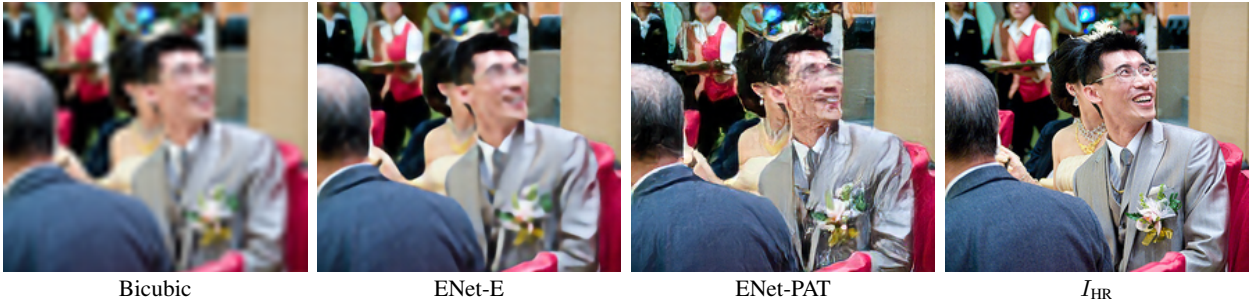

Figure 8. Failure case for ENet-PAT on an image from ImageNet at 4x super-resolution. While producing an overall sharper image than ENet-E, ENet-PAT fails to reproduce a realistically looking face, leading to a perceptually implausible result.

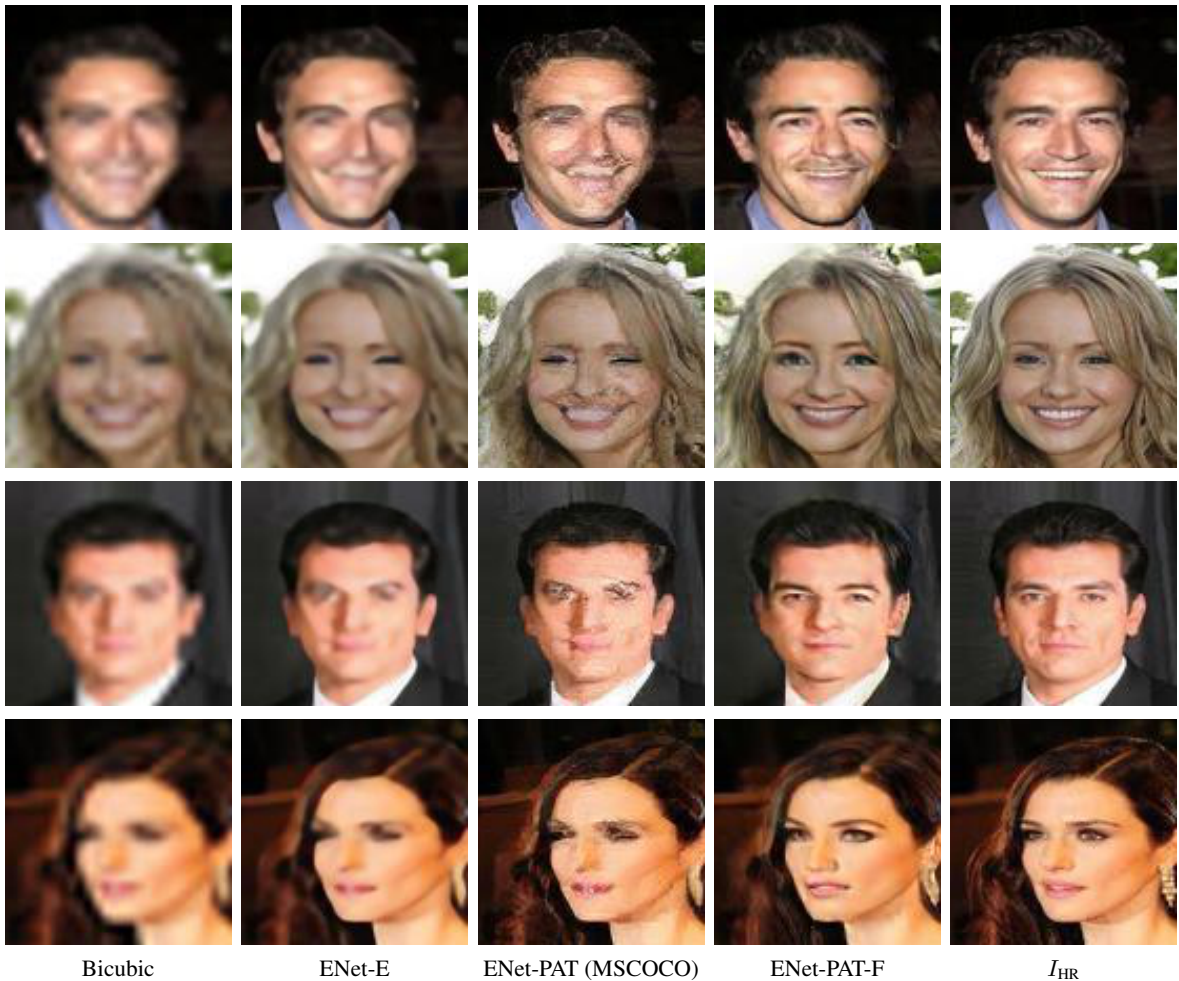

Figure 9. Comparing our models on images of faces at 4x super resolution. ENet-PAT produces artifacts since its training dataset did not contain many high-resolution images of faces. When trained specifically on a dataset of faces (ENet-PAT-F), the same network produces realistic very realistic images, though the results look different from the actual ground truth images (similar to the results in Yu and Porikli [20]). Note that we did not fine-tune the parameters of the losses for this specific task so better results may be possible.